\pdfoutput=1

\documentclass[11pt]{article}

\usepackage{emnlp2021}

\usepackage{times}
\usepackage{latexsym}

\usepackage[T1]{fontenc}

\usepackage[utf8]{inputenc}

\usepackage{microtype}

\usepackage{graphicx}
\usepackage{enumitem}
\usepackage{blindtext}

\newcommand*{\affaddr}[1]{#1} 
\newcommand*{\affmark}[1][*]{\textsuperscript{#1}}
\newcommand*{\email}[1]{\texttt{#1}}

\title{Augmenting BERT-style Models with Predictive Coding to Improve Discourse-level Representations}

\author{%
Vladimir Araujo\affmark[1,2], Andrés Villa\affmark[1], Marcelo Mendoza\affmark[3], Marie-Francine Moens\affmark[2], Alvaro Soto\affmark[1]\\
\affaddr{\affmark[1]Pontificia Universidad Católica de Chile},
\affaddr{\affmark[2]KU Leuven},\\
\affaddr{\affmark[3]Universidad Técnica Federico Santa María}\\
\email{\{vgaraujo,afvilla\}@uc.cl},
\email{mmendoza@inf.utfsm.cl}\\
\email{sien.moens@kuleuven.be},
\email{asoto@ing.puc.cl}
}

\begin{document}
\maketitle
\begin{abstract}
Current language models are usually trained using a self-supervised scheme, where the main focus is learning representations at the word or sentence level. However, there has been limited progress in generating useful discourse-level representations. In this work, we propose to use ideas from predictive coding theory to augment BERT-style language models with a mechanism that allows them to learn suitable discourse-level representations. As a result, our proposed approach is able to predict future sentences using explicit top-down connections that operate at the intermediate layers of the network. By experimenting with benchmarks designed to evaluate discourse-related knowledge using pre-trained sentence representations, we demonstrate that our approach improves performance in 6 out of 11 tasks by excelling in discourse relationship detection.

\end{abstract}

\vspace{-0.1cm}
\section{Introduction}
\vspace{-0.1cm}

Pre-trained language models are among the leading methods to learn useful representations for textual data. Several pre-training objectives have been proposed in recent years, such as causal language modeling \citep{radford2018,radford2019}, masked language modeling \citep{devlin-etal-2019-bert}, and permutation language modeling \citep{NEURIPS2019_dc6a7e65}. However, these approaches do not produce suitable representations at the discourse level \citep{huber-etal-2020-sentence}.

Simultaneously, neuroscience studies have suggested that predictive coding (PC) plays an essential role in language development in humans \citep{Ylinen2016,Zettersten2019}. PC postulates that the brain is continually making predictions of incoming sensory stimuli \citep{Rao1999,Friston2005,Clark2013,Hohwy2013}, with word prediction being the main mechanism \citep{VanBerkum2005,Kuperberg2015}. However, recent studies speculate that the predictive process could occur within and across utterances, fostering discourse comprehension \citep{Kandylaki2016,Pickering2018}.

In this work, we propose to extend BERT-type models with recursive bottom-up and top-down computation based on PC theory. Specifically, we incorporate top-down connections that, according to PC, convey predictions from upper to lower layers, which are contrasted with bottom-up representations to generate an error signal that is used to guide the optimization of the model. Using this approach, we attempt to build feature representations that capture discourse-level relationships by continually predicting future sentences in a latent space. We evaluate our approach on DiscoEval \citep{chen-etal-2019-evaluation} and SciDTB for discourse evaluation \citep{huber-etal-2020-sentence} to assess whether the embeddings produced by our model capture discourse properties of sentences without finetuning. 
Our model achieves competitive performance compared to baselines, especially in tasks that require to discover discourse-level relations.

\vspace{-0.1cm}
\section{Related Work}
\vspace{-0.1cm}

\subsection{BERT for Sentence Representation}
\vspace{-0.1cm}

Pre-trained self-supervised language models have become popular in recent years. BERT \citep{devlin-etal-2019-bert} adopts a transformer encoder using a masked language modeling (MLM) objective for word representation. It also proposes an additional loss called next-sentence prediction (NSP) to train a model that understands sentence relationships. On the other hand, ALBERT \citep{Lan2020ALBERT} proposes a loss based primarily on coherence called sentence-order prediction (SOP).

SBERT \citep{reimers-gurevych-2019-sentence} uses a siamese structure to obtain semantically meaningful sentence embeddings, focusing on textual similarity tasks. 
ConveRT \citep{henderson-etal-2020-convert} uses a dual-encoder to improve sentence embeddings for response selection tasks. These models focus on obtaining better representations for specialized sentence pair tasks, so they are not comparable with our which intended to be general-purpose.

More recently, SLM \citep{lee-etal-2020-slm} proposes a sentence unshuffling approach for a fine understanding of the relations among the sentences at the discourse level. CONPONO \citep{iter-etal-2020-pretraining} considers a discourse-level objective to predict the surrounding sentences given an anchor text. This work is related to our approach; the key difference is that our model predicts future sentences sequentially using a top-down pathway. We consider CONPONO as our main baseline.

\vspace{-0.1cm}
\subsection{Predictive Coding and Deep Learning}
\vspace{-0.1cm}

Recent work in computer vision takes inspiration from PC theory to build models for accurate \citep{NEURIPS2018_1c63926e} and robust \citep{NEURIPS2020_0660895c} image classification. PredNet \citep{DBLP:conf/iclr/LotterKC17} proposes a network capable of predicting future frames in a video sequence by making local predictions at each level using top-down connections. CPC \citep{oord2018representation} is an unsupervised learning approach to extract useful representations by predicting text in a latent space. Our method takes inspiration from these models, considering top-down connections and predictive processing in a latent space.

\vspace{-0.1cm}
\section{Proposed Method}
\label{sec:length}
\vspace{-0.1cm}   


\subsection{Model Details}
\vspace{-0.1cm}

\begin{figure}
  \includegraphics[width=0.47\textwidth]{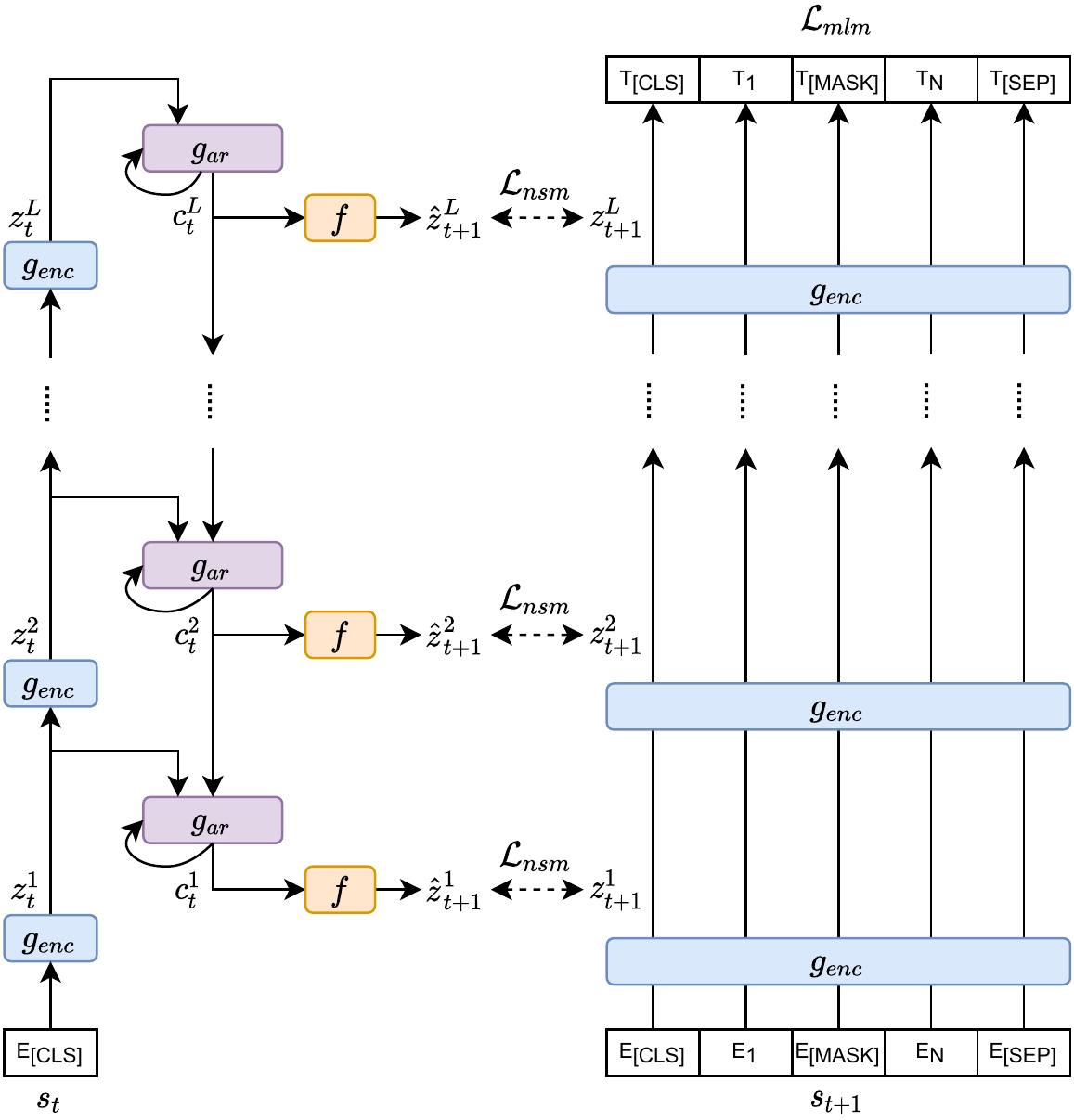}
  \vspace{-0.25cm}
  \caption{Example of prediction of one future sentence.
  Given an input sentence $s_t$ at time step $t$, the corresponding representation $z_{t}^{l}$ is calculated  at layer $l$. Then a context vector $c_{t}^{l}$ is computed via a top-down pathway (left). Afterwards, a future sentence $\hat{z}_{t + 1}^l$ is predicted to be compared to the actual representation ${z}_{t+1}$ (right).}
  \label{fig:predbert}
  \vspace{-0.25cm}   
\end{figure}

Our model consists of a BERT-style model as a sentence encoder (ALBERT and BERT are used in this work) and a GRU model \citep{cho-etal-2014-learning} that predicts next sentences (see Figure~\ref{fig:predbert}). 
Our intuition is that by giving the model the ability to predict future sentences using a top-down pathway, it will learn better relationships between sentences, thus improving sentence-level representations of each layer for downstream tasks.

The input is a sequence $s_1, s_2,.. , s_n$ of sentences extracted from a paragraph. We encode sentence $s_t$ with  encoder $g_{enc}$ that generates output $z_{t}^{l}$ at time step $t$ and layer $l$ ($l$ is from $1$ to $L$). Note that vector $z_{t}^{l}$ is obtained from the special token [CLS], which is commonly used as sentence representation. Next, an autoregressive model $g_{ar}$ produces a context vector $c_{t}^{l}$ as a function of $z_{t}^{l}$ and the context vectors of the upper layer $c_{t}^{l+1}$ and the previous step $c_{t-1}^{l}$. 

\vspace{-0.5cm}
\begin{equation}\label{eq:encoder}
z_{t}^{l}  = g_{enc}(s_t),\: c_{t}^{l} = g_{ar}(z_{t}^{l}, c_{t} ^{l+1}, c_{t-1}^{l})
\end{equation}

Then we introduce a predictive function $f(.)$ to predict a future sentence. In other words, $f(.)$ takes as input the context representation $c_t^l$ from time step $t$ and layer $l$, and predicts the latent representation $\hat{z}_{t+1}^l$ at time step $t+1$, \textit{i.e.}: 

\vspace{-0.4cm}
\begin{equation}\label{eq:predictive}
\hat{z}_{t+1}^{l} = f(c_{t}^{l})
\end{equation} 

In the spirit of Seq2Seq \citep{NIPS2014_a14ac55a}, representations are predicted sequentially, which differs from the CONPONO model that predicts $k$ future sentences with a unique context vector.

\vspace{-0.1cm}
\subsection{Loss Function}
\vspace{-0.1cm}

We rely on the InfoNCE loss proposed for the CPC model \cite{oord2018representation}. 
This constructs a binary task in which the 
goal is to classify one real sample among many noise samples. InfoNCE encourages the predicted representation $\hat{z}$ to be close to the ground truth $z$.

In the forward pass, the ground truth representation $z$ and the predicted representation $\hat{z}$ are computed at each layer of the model. So we denote the corresponding feature vectors as $z_{i}^{j}$ and $\hat{z}_{i}^{j}$ where $i$ denotes the temporal index and $j$ is the layer index. A dot product computes the similarity between the predicted and ground truth pair. Then, we optimize a cross-entropy loss that distinguishes the positive pair out of all other negative pairs:


\vspace{-0.5cm}
\begin{equation}\label{eq:objective}
\mathcal{L}_{nsm} = -\sum_{i,j} \left[ \log \frac{ \exp(\hat{z}_{i}^{{j}^\top} \cdot z_{i}^{j}) } { \sum_{m} \exp(\hat{z}_{i}^{{j}^\top} \cdot z_{m}^{j}) } \right]
\end{equation}


There is only one positive pair ($\hat{z}_{i}^{j}$, $z_{i}^{j}$) for a predicted sentence $\hat{z}_{i}^{j}$, which are the features at the same time step and the same layer. The rest of pairs ($\hat{z}_{i}^{j}$, $z_{m}^{j}$) are negative pairs, where ($i,j$) $\neq$  ($m,j$). In practice, we draw negative samples from the batch. This is a simple method and a more complex generation of negative samples could improve results. Our loss function, which we refer to as next-sentence modeling ($\mathcal {L}_{nsm}$), is used in conjunction with the BERT masked language model loss ($\mathcal {L}_{mlm}$). Accordingly, we train to minimize:

\vspace{-0.2cm}
\begin{equation}\label{eq:sumlosses}
\mathcal{L} = \mathcal{L}_{nsm} + \mathcal{L}_{mlm}
\end{equation} 
\vspace{-0.5cm}


\vspace{-0.1cm}
\subsection{Pre-training and Implementation Details}
\vspace{-0.1cm}

We extend ALBERT and BERT models, obtaining PredALBERT and PredBERT as a result. As mentioned above, our models are fed with a set of contiguous sentences $s_n$ that are processed one-at-a-time.
Note that the length of the conventional BERT input is 512 tokens. However, it is unlikely that a sentence will have that many tokens. 

We join 3 contiguous sentences to create a long sequence. Longer sequences are truncated, and shorter sequences are padded.
We use an overlapping sentence between contiguous sentence groups. For instance, given a paragraph $s_1, s_2,.. , s_{9}$, the 1st sequence is $s_1, s_2, s_3$, the 2nd sequence is $s_3, s_4, s_5$, and so on.
Our early experiments show that this setting improves the model's predictive ability in the validation set.
We hypothesize that the model can predict up to 3 sentences by using information from the overlapping sentences.

We pre-train our models with the predictive mechanism set to predict the next 2 future sentences ($k=2$). At time 1, our model represents sequence 1, then this vector feeds the top-down flow (GRU) generating a context representation in each layer that is used to predict sequence 2. Then, the model represents sequence 2 to contrast it with the predicted one. This is repeated one more time to reach $k=2$ predicted future sequences. For a fair comparison, we train using the BookCorpus \citep{7410368} and Wikipedia datasets, as well as the BERT, ALBERT, and CONPONO models.

Note that top-down connections are only available during pre-training.
At evaluation time, we discard the top-down connections keeping only the encoder, thus obtaining a model equivalent to BERT or ALBERT in terms of the parameters. Table~\ref{table:params} shows the number of parameters in our models.

\begin{table}[ht]
\centering
\setlength\tabcolsep{1.25mm}
\begin{tabular}{l|cc|c|}
\cline{2-4}
                                     & \multicolumn{2}{c|}{\textbf{Parameters}} & \multicolumn{1}{l|}{\textbf{Layers}} \\ \cline{1-3}
\multicolumn{1}{|l|}{\textbf{Model}} & \textbf{Training}  & \textbf{Inference}  & \multicolumn{1}{l|}{}                                 \\ \hline
\multicolumn{1}{|l|}{PredALBERT-B}   & 17.7M              & 11.7M               & 12                                                    \\
\multicolumn{1}{|l|}{PredALBERT-L}   & 28.3M              & 17.7M               & 24                                                    \\
\multicolumn{1}{|l|}{PredBERT-B}     & 114.2M             & 109.5M              & 12                                                    \\ \hline
\end{tabular}
\vspace{-0.3cm}
\caption{\label{table:params} Number of parameters of PredBERT models.}
\vspace{-0.5cm}
\end{table}


We used the Huggingface library \citep{wolf-etal-2020-transformers} to implement our models.
We initialize the encoder model with BERT or ALBERT weights depending on the version.
The autoregressive model was initialized with random weights. 
For model efficiency in both versions, we use parameter-sharing across layers in the autoregressive model.
We trained the models for 1M steps with batch size 8. We use Adam optimizer with weight decay and learning rate of 5e-5.
For the masked language modeling, we consider dynamic masking, where the masking pattern is generated every time we feed a sequence to the model. Unlike BERT, we mask 10\% of all tokens in each sequence at random.


\vspace{-0.1cm}
\section{Experiments}
\vspace{-0.1cm}

\subsection{Datasets}
\vspace{-0.1cm}

Our focus is to evaluate if the discourse properties of sentences are captured by our model without finetuning.
DiscoEval \citep{chen-etal-2019-evaluation} and SciDTB \citep{huber-etal-2020-sentence} datasets include probing tasks designed for discourse evaluation, thus letting us know what discourse-related knowledge our model is capturing effectively.

\paragraph{DiscoEval:} Suite of tasks to evaluate discourse-related knowledge in sentence representations.
It includes 7 tasks: Sentence position (SP), Binary sentence ordering (BSO), Discourse coherence (DC), Sentence section prediction (SSP), Penn discourse tree bank (PDTB-E/I), and Rhetorical structure theory (RST). SP, BSO, DC, and SSP assess discourse coherence with binary classification, while PDTB and RST assess discourse relations between sentences through multi-class classification.

\paragraph{SciDTB-DE:} Set of tasks designed to determine whether an encoder captures discourse properties from scientific texts. It considers 4 tasks: Swapped units detection (Swapped), Scrambled sentence detection (Scrambled), Relation detection (BinRel), and Relation semantics detection (SemRel). Both Swapped and Scrambled tasks were designed for clause coherence verification, while BinRel and SemRel for discourse relationship detection. 

\begin{table*}[!ht]
\centering
\vspace{-0.5cm}
\setlength\tabcolsep{1.25mm}
\begin{tabular}{l|cccccc|cccc|}
\cline{2-11}
                                     & \multicolumn{6}{c|}{\textbf{DiscoEval}}                                                                                                                                        & \multicolumn{4}{c|}{\textbf{SciDTB-DE}}                             \\ \hline
\multicolumn{1}{|l|}{\textbf{Model}} & \textbf{SP}               & \textbf{BSO}              & \textbf{DC}               & \textbf{SSP}              & \textbf{PDTB-E/I}                 & \textbf{RST}               & \textbf{BinRel} & \textbf{SemRel} & \textbf{Swap}  & \textbf{Scram} \\ \hline
\multicolumn{1}{|l|}{ALBERT-B}       & 52.11                     & 70.10                     & 53.89                     & 80.10                     & 37.10 / 39.28                     & 54.03                      & 71.48           & 56.32           & 92.04          & 84.57          \\
\multicolumn{1}{|l|}{ALBERT-L}       & 52.61                     & 70.91                     & 53.85                     & 80.45                     & 37.44 / 38.15                     & 54.71                      & 72.03           & 54.86           & \textbf{92.67} & \textbf{87.79} \\
\multicolumn{1}{|l|}{BERT-B}         & 53.80                     & 72.33                     & 59.27                     & 80.37                     & 42.52 / 41.97                     & 59.63                      & 76.17           & 64.93           & 92.07          & 86.91          \\
\multicolumn{1}{|l|}{CONPONO}        & \textbf{54.52}            & \textbf{72.81}            & 58.62                     & \textbf{80.87}            & 41.37 / 41.27                     & 59.74                      & 77.86           & 68.75           & 91.76          & 83.62          \\ \hline
\multicolumn{1}{|l|}{PredALBERT-B}   & \multicolumn{1}{l}{50.59} & \multicolumn{1}{l}{69.64} & \multicolumn{1}{l}{61.69} & \multicolumn{1}{l}{79.60} & \multicolumn{1}{l}{40.71 / 41.40} & \multicolumn{1}{l|}{58.54} & 77.83           & 65.49           & 92.51          & 83.38          \\
\multicolumn{1}{|l|}{PredALBERT-L}   & 51.70                     & 70.35                     & 61.87                     & 79.85                     & \textbf{42.66 / 42.92}            & \textbf{59.93}             & \textbf{78.11}  & 70.15           & 92.67          & 84.79          \\
\multicolumn{1}{|l|}{PredBERT-B}     & 50.80                     & 69.94                     & \textbf{62.25}            & 79.83                     & 40.10 / 42.20                     & 59.58                      & 76.21           & \textbf{72.92}  & 91.06          & 80.18          \\ \hline
\end{tabular}
\vspace{-0.3cm}
\caption{\label{table:results} Accuracy results in the DiscoEval and SciDTB-DE datasets. 
We carry out the evaluation 10 times with different seeds and report the average across the trials.
B and L indicate the base and large versions, respectively.
}
\vspace{-0.5cm}
\end{table*}

\vspace{-0.1cm}
\subsection{Experimental Setup}
\label{section:experimental}

\paragraph{Baselines:}
Following \citet{chen-etal-2019-evaluation,huber-etal-2020-sentence}, we include the results of BERT Base \citep{devlin-etal-2019-bert}. We also evaluate CONPONO \citep{iter-etal-2020-pretraining}, which is the most related model to our approach. 
Because these models have more parameters than PredBERT, we also include ALBERT \citep{Lan2020ALBERT} Base and Large, which are directly comparable to our model.
For a fair and consistent comparison, we rerun all baseline evaluations. We use the pre-trained Huggingface models \citep{wolf-etal-2020-transformers} for BERT and ALBERT. In the case of CONPONO, we use a version pre-trained to predict 2 next surrounding sentences\footnote{\href{https://github.com/google-research/language/tree/master/language/conpono}{https://github.com/google-research/language/conpono}}.

\paragraph{Evaluation:} In the case of DiscoEval, we use the original code provided by \citet{chen-etal-2019-evaluation}. 
We observe that this configuration leads to CONPONO model results that differ from the reported on the original paper.
On the other hand, following \citet{huber-etal-2020-sentence}, we use SentEval \citep{conneau-kiela-2018-senteval} toolkit for SciDTB-DE evaluation.
In both cases, the process involves loading a pre-trained model with frozen weights and training a logistic regression on top of the sentence embeddings. 
To train, we use the average of sentence representations ([CLS]) from all the layers.

\vspace{-0.1cm}
\section{Results}
\vspace{-0.1cm}

Table~\ref{table:results} shows the results of our models.
We observe improvements in discourse relation detection (PDTB, RST, BinRel, SemRel) and discourse coherence (DC) tasks compared to the best baseline (CONPONO).
Across these tasks, PredALBERT-L outperforms by $\sim$1.34 points on average, while PredBERT-B by $\sim$0.94.
PredALBERT-B achieves competitive performance but does not outperform CONPONO.
However, if we compare our models with their direct baselines (ALBERT-B/L, BERT-B), the increase is greater. PredALBERT-B by $\sim$3.92, PredALBERT-L by $\sim$7.43, and PredBERT-B by $\sim$1.46 points on average. 
The Stuart–Maxwell tests demonstrated a significant difference between our best model PredALBERT-L and ALBERT-L ($p = 0.009$) or CONPONO ($p = 0.05$). 
We also highlight that PredALBERT-B/L achieves competitive performance with fewer parameters than BERT and CONPONO.


Decreased performance of our models in the SP, BSO, SSP, Swap, and Scram tasks is due to the fact they are closely related to the baselines optimization objectives, which consist of sentence order prediction (ALBERT), topic prediction (BERT), or a combination of them (CONPONO). 
In contrast, our approach uses a next sentence prediction task in a generative way that encourages the capture of discourse relationships, improving its performance on PDTB, RST, BinRel, and SemRel tasks.
%


\vspace{-0.1cm}
\section{Ablation Study}
\vspace{-0.1cm}

\begin{table*}[ht]
\centering
\setlength\tabcolsep{0.75mm}
\begin{tabular}{l|cccccc|c|cccc|c|}
\cline{2-13}
                                     & \multicolumn{7}{c|}{\textbf{DiscoEval}}                                                              & \multicolumn{5}{c|}{\textbf{SciDTB-DE}}                            \\ \hline
\multicolumn{1}{|l|}{\textbf{Model}} & \textbf{SP} & \textbf{BSO} & \textbf{DC}    & \textbf{SSP} & \textbf{PDTB-E/I}      & \textbf{RST}   & {Avg}   & \textbf{BinRel} & \textbf{SemRel} & \textbf{Swap} & \textbf{Scram} & {Avg} \\ \hline
\multicolumn{1}{|l|}{\texttt{Default}} & 50.59 & 69.64 & 61.69 & 79.60 & 40.71 / 41.40 & 58.54 & \underline{57.45} & 77.83 & 65.49 & 92.51 & 83.38 & \underline{79.80} \\
\multicolumn{1}{|l|}{\texttt{Half}} & 51.43 & 70.03 & 62.71 & 79.88 & 40.26 / 40.67 & 57.37 & \underline{57.48} & 79.73  & 63.54  & 91.54 & 84.95 & \underline{79.94} \\
\multicolumn{1}{|l|}{\texttt{Last}} & 49.73 & 68.36 & 62.88 & 79.33 & 38.80 / 39.62 & 57.06 & \underline{56.54} & 74.00 & 59.51 & 92.64 & 83.42 & \underline{77.39} \\
\multicolumn{1}{|l|}{\texttt{w/o TDC}} & 45.67 & 65.76 & 57.94 & 78.38 & 37.49 / 39.32 & 57.37 & \underline{54.56} & 72.14 & 60.55 & 88.81 & 79.99 & \underline{75.37} \\ \hline
\end{tabular}
\vspace{-0.3cm}
\caption{\label{table:fullablation} Results of ablation experiments with PredALBERT-B as the default model.}
\vspace{-0.3cm}
\end{table*}

In order to verify the influence of PC mechanism on the pre-training result, we carry out ablation experiments. 
We use our PredALBERT-B as the \texttt{Default} model, which includes top-down connections and recurrence from layer 12 to layer 1. 
Ablations involve removing top-down connections and the recurrence of certain layers. Table~\ref{table:fullablation} shows performance across all tasks for each benchmark.

The first experiment uses the PC mechanism on \texttt{Half} the layers, \textit{i.e.}, the GRU and predictive layer are present from layer 12 to layer 6. This variation exceeds the \texttt{Default} model by $\sim$0.03 in DiscoEval and $\sim$0.14 in SciDTB-DE.
The second experiment uses the PC mechanism only on the \texttt{Last} layer of the transformer. It means that the combination of the GRU and prediction layer is only present in layer 12. This reduces the performance by $\sim$0.91 in DiscoEval and $\sim$2.41 in SciDTB-DE. 




Also, we conducted an additional experiment where we removed the top-down connections ($\texttt{w/o TDC}$) to the \texttt{Default} model. This is equivalent to modifying equation~\ref{eq:encoder} by $g_{ar}(z_t^l,c_{t-1}^l)$. We found that this ablation severely affects the performance of the \texttt{Default} model performance by $\sim$2.89 in DiscoEval and $\sim$4.43 in SciDTB-DE.



Our findings indicate that top-down pathway is beneficial for improving discourse representations of BERT-type models. However, it is not clear in which layers it is crucial to have the PC mechanism. We hypothesize that this is related to the fact that the BERT-style models encode syntactic and semantic features in different layers \citep{jawahar-etal-2019-bert,aspillaga-etal-2021-inspecting}, so a specialized PC mechanism for syntax or semantics would be desirable. We left this study for future work.


\section{What Does The Model Learn?}
\vspace{-0.2cm}

Because our model excels at detecting discourse relations, in this section, we explore whether the resulting vectors actually represent the role of a sentence in its discursive context.
To illustrate what PredBERT learns, we follow the methodology proposed by \citet{lee-etal-2020-slm}. 
Using cosine similarity, we use labeled sentences with discourse relations as queries to retrieve the top 3 most similar sentences from an unlabeled corpus using.
We use sentences from the MIT Discourse Relations Annotation Guide\footnote{\href{http://projects.csail.mit.edu/workbench/update/guides/10\%20-\%20Discourse\%20Relations.pdf}{https://bit.ly/3z45IG2}} as queries and the Gutenberg dataset \citep{lahiri-2014-complexity} as the unlabeled corpus.
We compute the representations as mentioned in Section~\ref{section:experimental}.
This process allows us to identify if similar vectors share the same or equivalent discourse relations.



\paragraph{Temporal relation:}  \textbf{Query} = He \textcolor{magenta}{knows} a \textcolor{blue}{tasty meal} \textcolor{red}{when} he eats one.
\begin{enumerate}[noitemsep,topsep=0pt]
\item The last five words took Tuppence's fancy mightily, especially \textcolor{red}{after} a \textcolor{blue}{meagre breakfast} and a supper of buns the night before.
\item I \textcolor{magenta}{know} a disinterested man \textcolor{red}{when} I see him.
\item He had about ten pounds \textcolor{red}{when} I found him.
\end{enumerate}

Sentence 1 has a \textit{succession} relation due to the use of the word \textcolor{red}{after}. Sentence 3 shows a \textit{synchrony} relation because it uses \textcolor{red}{when} as the query. 
Sentence 2 does not have a \textit{temporal} relation.

\paragraph{Comparison relation:}  \textbf{Query} = IBM’s \textcolor{blue}{stock price} rose, \textcolor{red}{but} the overall \textcolor{magenta}{market} fell.
\begin{enumerate}[noitemsep,topsep=0pt]
\item The \textcolor{blue}{stock} \textcolor{magenta}{markets} of the world gambled upon its chances, and its bonds at one time were high in favor.
\item Tommy's heart beat faster, \textcolor{red}{but} his casual pleasantness did not waver.
\item I guess I was just a mite hasty, \textcolor{red}{but} I've been feeling bad about this \textcolor{blue}{money} question.
\end{enumerate}

Sentence 1 matched \textcolor{blue}{stock} and \textcolor{magenta}{market} words but does not contain a \textit{comparison} relation. Sentences 2 and 3 include a \textit{counter-expectation} relation similar to the query that uses the word \textcolor{red}{but}.

\paragraph{Contingency relation:}  \textbf{Query} = I refused to \textcolor{blue}{pay} the cobbler the full \$95 \textcolor{red}{because} he did \textcolor{magenta}{poor} \textcolor{blue}{work}.
\begin{enumerate}[noitemsep,topsep=0pt]
\item I did the labor of writing one address this year, \textcolor{red}{and} got thunder for my \textcolor{blue}{reward}.
\item I don't believe in a law to prevent a man from getting \textcolor{magenta}{rich}; it would do more harm than good.
\item \textcolor{red}{When} I fixed a plan for an election in Arkansas I did it in ignorance that your convention was doing the same \textcolor{blue}{work}.
\end{enumerate}

All sentences contain semantically related words like \textcolor{blue}{pay/reward} and \textcolor{magenta}{poor/rich}. Sentences 1 and 2 include a \textit{cause} relation explicit and implicit; this is related to a query that has \textit{pragmatic cause} relation.
However, sentence 3 shows a \textit{temporal} relation.

\vspace{-0.1cm}
\section{Conclusions}
\vspace{-0.1cm}

We introduce an approach based on PC theory, which extends BERT-style models with recursive bottom-up and top-down computation along with a discourse representation objective.
Our models achieve competitive results in discourse analysis tasks, excelling in relations detection. 

\section*{Acknowledgements}
This work was supported in part by the Millennium Institute for Foundational Research on Data (IMFD), the European Research Council Advanced Grant 788506 and the TPU Research Cloud (TRC) program.
The first author is grateful to Rodrigo Tufiño for helpful technical advice.

\bibliography{anthology,custom}
\bibliographystyle{acl_natbib}

\newpage

\end{document}